\documentclass{article}
\usepackage{ijcai20}

\usepackage{times}
\usepackage{soul}
\usepackage{url}
\usepackage[hidelinks]{hyperref}
\usepackage[utf8]{inputenc}
\usepackage[small]{caption}
\usepackage{graphicx}
\usepackage{amsmath,amssymb}
\usepackage{amsthm}
\usepackage{booktabs}
\usepackage{algorithm}
\usepackage{algorithmic}
\usepackage{subcaption}
\usepackage{caption}

\usepackage{diagbox}
\newcommand{\tabincell}[2]{\begin{tabular}{@{}#1@{}}#2\end{tabular}}

\newcommand{\PHB}[1]{\noindent\textbf{#1}\hspace{.5em}} % paragraph heading at the beginning of a section
\newcommand{\PHM}[1]{\vspace{.4em}\noindent\textbf{#1}\hspace{.5em}} % paragraph heading in the middle of a section

\newtheorem{theorem}{Theorem}

\title{Learning to Detect Malicious Clients for Robust Federated Learning}

\author{
Suyi Li$^1$\and
Yong Cheng$^2$\and
Wei Wang$^1$\\
Yang Liu$^2$\and
Tianjian Chen$^2$\\
\affiliations
$^1$The Hong Kong University of Science and Technology\\
$^2$AI Department, WeBank\\
\emails
\{slida, weiwa\}@cse.ust.hk,
\{petercheng, yangliu, tobychen\}@webank.com
}

\begin{document}

\maketitle

\begin{abstract}
Federated learning systems are vulnerable to  attacks from  malicious  clients. As the central server in the system cannot govern the behaviors of the clients,  a rogue client may initiate an attack by sending malicious model updates to the server, so as to degrade   the learning performance or enforce  targeted model poisoning attacks (a.k.a. backdoor attacks).   Therefore, timely detecting these malicious model updates and the underlying attackers becomes critically important. In this work, we propose a new framework for robust federated learning where the central server learns to \textit{detect and remove} the malicious model updates using a powerful detection model, leading to \textit{targeted defense}.   We evaluate our solution in both image classification and sentiment analysis tasks with a variety of machine learning models. Experimental results show that our solution ensures robust federated learning that is resilient to both the Byzantine attacks and the targeted model poisoning attacks.
\end{abstract}

\section{Introduction}
\label{sec:intro}

Federated learning (FL) comes as a new distributed machine learning (ML) paradigm  where multiple clients (e.g., mobile  devices) collaboratively train an ML model without
revealing their private
data~\cite{pmlr-v54-mcmahan17a,QiangYangTIST2019,kairouz2019advances}. In a
typical FL setting, a central server is used to maintain a global model and
coordinate the clients. Each client transfers the local model updates to the
central server for immediate aggregation, while keeping the raw data in their 
local storage.  As no private data gets exchanged in the training process, 
FL provides a strong privacy guarantee to the participating clients and has found wide applications in edge computing, finance, and
healthcare~\cite{YangBookFL2019,BingshengHe2019Overview,li2019federated}. 

%Despite the benefits of privacy preservation, 
FL systems are vulnerable to
attacks from malicious clients, which has become a major roadblock to
their practical deployment~\cite{bhagoji2018analyzing,bagdasaryan2018backdoor,wu2019federated,kairouz2019advances}.  In an FL system, the central server cannot govern
the behaviors of the clients, nor can it access their private data. As a
consequence, the malicious clients can cheat the server by sending modified and 
\emph{harmful} model updates, initiating \emph{adversarial attacks} on the
global  model~\cite{kairouz2019advances}. In this paper, we consider two types
of adversarial attacks, namely the \emph{untargeted} attacks and the
\emph{targeted} attacks. The untargeted attacks aim to degrade the overall
model performance and can be viewed as Byzantine attacks which result in model
performance deterioration or failure of model training~\cite{LipingLi2019,wu2019federated}. The targeted attacks (a.k.a. backdoor attacks)~\cite{bhagoji2018analyzing,bagdasaryan2018backdoor,sun2019can}, on
the other hand, aim to modify the behaviors of the model on some specific data
instances chosen by the attackers (e.g., recognizing the images of cats as
dogs), while keeping the model performance on the other data instances
unaffected. Both the untargeted and targeted attacks can result in
catastrophic consequences. Therefore, attackers, along with their harmful
model updates, must be timely \textit{detected and removed}  from an FL system to
prevent malicious model corruptions and inappropriate incentive awards
distributed to the adversary
clients~\cite{Incentive2019}.

%There is a rich body of work on 
Defending against Byzantine attacks has been extensively studied  in
distributed ML,
e.g.,~\cite{chen2017distributed,NIPS2017_6617,xie2018generalized,pmlr-v80-yin18a}.
However, we find that the existing Byzantine-tolerant algorithms are unable to
achieve satisfactory model performance in the FL setting. These methods do not
differentiate the malicious updates from the normal ones. Instead, they aim to
tolerate the adversarial attacks and mitigate their negative impacts with new
model update mechanisms that cannot be easily compromised by the
attackers. In addition, most of these methods assume independent and identically distributed~(IID) data, making them a poor fit
in the FL scenario where non-IID  datasets   are commonplace. Researchers in the FL community have also proposed
various defense mechanisms against adversarial
attacks~\cite{sun2019can,Auror2016}. These mechanisms, however,
are mainly designed for the deliberate targeted attacks and cannot survive under the
untargeted Byzantine attacks.

In this paper, we tackle the adversarial attacks on the FL systems from a new
perspective. We propose a \textit{spectral anomaly detection} based
framework~\cite{chandola2009anomaly,kieu2019outlier,an2015variational} that
detects the abnormal model updates based on their \textit{low-dimensional
embeddings}, in which the noisy and  irrelevant features are removed whilst the
essential features are retained. We  show that in such a
low-dimensional latent feature space, the abnormal (i.e., malicious)  model  updates from clients  can be easily differentiated
as their essential features are drastically different from those of the normal updates, leading to \textit{ targeted defense}. 

To our best knowledge, we are the first to employ spectral anomaly detection for robust FL  systems.  Our spectral anomaly detection framework  provides three benefits. \emph{First},
it works in both the unsupervised and semi-supervised settings, making it
particularly attractive to the FL scenarios in which the malicious model
updates are unknown and cannot be accurately predicted beforehand.
\emph{Second}, our spectral anomaly detection model uses variational
autoencoder (VAE) with \emph{dynamic thresholding}. Because the detection
threshold is only determined after the model updates from all the clients
have been received, the attackers cannot learn the detection mechanism \emph{a
priori}. \emph{Third}, by detecting and removing the malicious updates in the
central server, their negative impacts can be fully eliminated. 

% \emph{Fourth},
% our analysis on the adversarial attacks shows that the impact of malicious
% updates from the untargeted attacks is positively related to their assigned
% weights in the model aggregation. 

% Our detection-based method leads to targeted defense against attacks in FL systems. 

We evaluate our spectral anomaly detection approach against the image
classification and sentiment analysis tasks in the heterogeneous FL settings
with various ML models, including logistic regression (LR), convolutional
neural network (CNN), and recurrent neural network
(RNN)~\cite{zhang2020dive}.  In all experiments, our method accurately
detects a range of adversarial attacks (untargeted and targeted) and
eliminates their negative impacts almost entirely. This is not possible using
the existing  Byzantine-tolerant approaches.

\section{Prior Arts}
\label{sec:related}

\subsection{Robust Distributed Machine Learning}
\label{sec:robust_ML}
%Developing robust distributed ML algorithms against Byzantine attacks has recently attracted a great amount of attentions. 
Existing  methods mainly focus
on building a \emph{robust  aggregator} that estimates the  ``center"
of the received local model updates rather than taking a weighted average,
which can be easily compromised. Most of these works assume IID data across all the clients (a.k.a. workers).  So the local model updates from any of the
benign  clients can presumably approximate the true gradients or model weights.
Following this idea, many robust ML algorithms, such as
Krum~\cite{NIPS2017_6617}, Medoid~\cite{xie2018generalized}, and Marginal
Median~\cite{xie2018generalized}, select a \emph{representative} client
and use its update to estimate the true center. This approach,
while statistically resilient to the adversarial attacks,
may result in  a \emph{biased} global model  as it only accounts for a small fraction of the local updates. 

Other approaches, such as GeoMed~\cite{chen2017distributed} and
Trimmed Mean~\cite{pmlr-v80-yin18a}, estimate the center based on the model updates
from clients, without differentiating the malicious from the normal
ones. These approaches can  mitigate the impacts of malicious attacks
to a certain degree but not fully eliminate them.

More recently, \cite{LipingLi2019}  introduces an additional     $l_{1}$-norm 
regularization on the cost function  to achieve robustness against   Byzantine
attacks in  distributed learning.   \cite{wu2019federated}  proposes an
approach that combines  distributed SAGA and geometric median   for robust 
federated optimization in the presence of Byzantine attacks. Both approaches
cannot defend against the targeted attacks.  

%There also exist distributed optimization based solutions that modify
%the procedure of updating local model weights, such
%as~\cite{LipingLi2019,wu2019federated}.
%However, these methods are only effective with strict assumptions and limited number of attackers\cite{chen2017distributed,pmlr-v80-yin18a}, which cannot be guaranteed in FL tasks. 

\subsection{Robust Federated Learning}
\label{sec:robust_FL}

The existing solutions for robust FL are mostly defense-based and are limited
to the targeted attacks. For example, \cite{Auror2016} proposes a
detection-based approach for backdoor attacks in collaborative ML.
However, it is assumed that the generated mask features of the training data
have the same distribution as that of the training data,
which is not the case in the FL setting. \cite{sun2019can} proposes a
low-complexity defense mechanism that mitigates the impact of backdoor attacks
in FL tasks through model weight clipping and noise injection. However, this
defense approach is unable to handle the untargeted attacks that do not modify
the magnitude of model weights, such as sign-flipping
attack~\cite{LipingLi2019}.  \cite{fang2019local} proposes two defense
mechanisms, namely error rate based rejection and loss function based
rejection, which sequentially reject the malicious local updates by
testing their impacts on the global model over a validation set. However, as FL
tasks typically involve a large number of clients, exhaustively testing their
impacts over the validation set is computationally prohibitive.

\subsection{Spectral Anomaly Detection}
\label{sec:spectral_anomaly_detection}

Spectral anomaly detection is one of  the most effective anomaly detection
approaches~\cite{chandola2009anomaly}. The idea is to embed both the normal
data instances and the abnormal instances into a low-dimensional  latent space (hence  the name  ``spectral"),  in
which their embeddings differ significantly. Therefore, by learning to remove
the noisy features of data instances and project the important ones into a
low-dimensional latent space, we can easily identify the abnormal
instances by looking at reconstruction errors~\cite{an2015variational}. This method has been proved effective in
detecting anomalous image data and time series
data~\cite{agovic2008anomaly,an2015variational,xu2018unsupervised,kieu2019outlier}.
%To our best knowledge, we are the first to employ spectral anomaly detection for robust federated learning systems. 

\section{Spectral Anomaly Detection  for Robust FL}
\label{sec:algorithm_design}

In this section, we present a novel spectral anomaly detection framework for
robust FL. 

\subsection{Problem Definition}
\label{sec:problem_def}

We consider a typical FL setting in which multiple clients collaboratively
train an ML model maintained in a central server using the \texttt{FedAvg}
algorithm~\cite{pmlr-v54-mcmahan17a}. We assume that an attacker can only
inspect a stale version of the model (i.e., \emph{stale whitebox} model
inspection~\cite{kairouz2019advances}), which is generally the case in FL. We
also assume the availability of a public  dataset that can be used for training
the spectral anomaly detection model. This assumption generally holds in
practice~\cite{li2019fedmd}. In fact, having a public dataset is indispensible
to the design of neural network architecture in FL. We defer the detailed
training process of the spectral anomaly detection model to
Section~\ref{subsec:data_model}.

\subsection{Impact of Malicious Model Updates}
\label{sec:impact_of_attack}

Before presenting our solution, we need to understand how the adversarial
updates may harm the model performance. To this end, we turn to a simple
linear model, where we quantify the negative impacts of those malicious
updates and draw key insights that drive our design. Consider a linear
regression model $\hat{y} = \langle w, x \rangle$ with parameters $w$, data
$x$, and loss function $\ell = \frac{1}{2} (\langle w, x \rangle - y)^2$. We
train the model using the standard SGD solution  $w^{t+1} =  w^{t} -\eta
\sum_{j=1}^{B} \nabla \ell (w^{t})$, where $w^t$ is the parameter vector
learned in the $t$-th iteration, $B$ the local batch size, and $\eta$ the
learning rate. Let $w^t_k$ be the model weight learned by the $k$-th client in
the $t$-th iteration without any malicious attacks. Let $\hat{w}^{t}_k$ be
similarly defined subject to attacks, where the malicious updates from the
adversarial clients are generated by adding noise $\psi$ to the normal
updates. The following theorem quantifies the negative impact of malicious updates. 

% \begin{theorem}[Impact of malicious model  updates] We have the following analytical result. 
% \begin{align}
% \label{eq:abnormal influence}
% \footnotesize
% \begin{split}
%     \mathbb{E}\left[w^{t + 1}_k \right] &= \mathbb{E}\left[ \varphi_{\eta}\left( s_n f^{t}+ s_a \cdot \psi \right) \right] \\
%     &= s_a \mathbb{E}[\psi] - s_a \mathbb{E}\left[\eta \sum_{j=1}^{B}\left(  \left\langle\psi, x_{(k, j)}\right\rangle x_{(k, j)}\right)\right] \\ 
%     &+ s_n \mathbb{E} \left[ f_k^{t+1} \right],
% \end{split}
% \end{align}
% \end{theorem}
\begin{theorem} 
\label{thm:negative_impact}
Let $f_a$ be the fraction of the total weights attributed to 
the malicious clients, where $0\le f_a \le 1$. We have 
\begin{equation}
\label{eq:abnormal_influence}
\footnotesize
\textstyle
    \mathbb{E}[\hat{w}^{t + 1}_k]  - \mathbb{E} [ w_k^{t+1}] = f_a ( \mathbb{E}[\psi] - \mathbb{E}\left[\eta \sum_{j=1}^{B} \langle\psi, x_{k, j}\rangle x_{k, j}\right]).
    % &+ \mathbb{E} \left[ f_k^{t+1} \right] ),
\end{equation}
\end{theorem}

\begin{figure}[tb]
\centering
\begin{subfigure}{0.045\linewidth}
    \includegraphics[width=\linewidth]{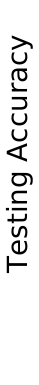}
\end{subfigure}
\begin{subfigure}{0.45\linewidth}
    \includegraphics[width=\linewidth]{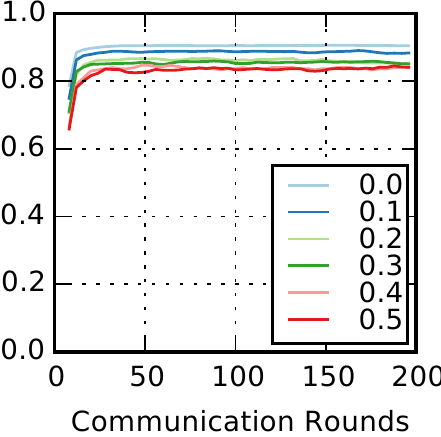}
    \caption{Additive noise attack.}
    \label{fig:mnist_noise_attack}
\end{subfigure}
\begin{subfigure}{0.45\linewidth}
    \includegraphics[width=\linewidth]{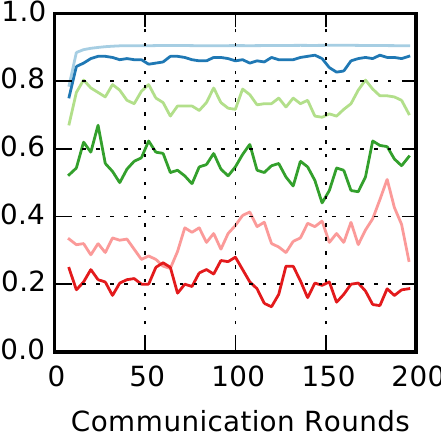}
    \caption{Sign-flipping attack.}
    \label{fig:mnist_sign_attack}
\end{subfigure}
\setlength{\abovecaptionskip}{2pt}
\caption{LR model accuracy. Curves  in the figure correspond to different sum of weights attributed to malicious attackers.}
\label{fig:mnist_different_fraction_attacker}
\end{figure}

We omit the proof of Theorem~\ref{thm:negative_impact} due to the space
constraint. Eq.~\eqref{eq:abnormal_influence} states that the impact of the
malicious updates is determined by two factors: (i) the noise $\psi$ added by
the attackers, and (ii) the fraction of total weights $f_a$ attributed to the
malicious clients in an FL system. We further confirm these observations with
simulation experiments shown in
Figure~\ref{fig:mnist_different_fraction_attacker}. With the same weights
attributed to the malicious clients in an FL system, sign-flipping attack
(Figure~\ref{fig:mnist_sign_attack}) can cause more significant damage on the
model performance than adding random noises
(Figure~\ref{fig:mnist_noise_attack}). Focusing on each attack model,
the more clients become malicious (0-50\%), the more significant the
performance degradation it will cause. 

Considering that the noise $\psi$ generated by the malicious clients is
unknown to the central server, the most effective way of eliminating the
malicious impact is to exclude their updates in model aggregation, i.e.,
setting $f_a$ to $0$. Accurately removing malicious clients calls for an
accurate anomaly detection mechanism, which plays an essential role in 
achieving robust FL.  

Eq.~\eqref{eq:abnormal_influence} also suggests that adding a small amount of
noise $\psi$ does not lead to a big deviation on the model weights. 
Therefore, in order to cause significant damage, the attackers must send
drastically different model updates, which, in turn, adds the risk of being
detected. Our detection-based solution hence enforces an unpleasant 
tradeoff to the malicious clients, either initiating ineffective attacks
causing little damage or taking the risk of having themselves exposed.

% As a result, we can set  $s_a$ to a sufficiently small value, so to \textit{detect  and then remove  the impact of the malicious clients}.  In this way,    we  can realize  robust FL for  scenarios   where  some of the clients are malicious and disguise themselves  as  benign  clients,  and   achieve  that have comparable performance to the case without malicious clients. 

\subsection{Malicious Clients Detection}

Following the intuitions drawn from a simple linear model, we propose to
detect the anomalous or malicious model updates in their low-dimensional
embeddings using spectral anomaly
detection~\cite{chandola2009anomaly,an2015variational,kieu2019outlier}.
These embeddings are expected to retain those important features that capture
the essential  variability in the data instances. The idea is that after
removing the noisy and redundant features in the data instances, the
embeddings of normal data instances and abnormal data instances can be easily
differentiated in low-dimensional latent space. One effective method to
approximate low-dimensional embeddings is to train a model with the
\emph{encoder-decoder} architecture. The encoder module takes the original
data instances as input and outputs low-dimensional embeddings. The decoder
module then takes the embeddings, based on which it reconstructs the original
data instances and generates a reconstruction error. The reconstruction error
is then used to optimize the parameters of the encoder-decoder model until it
converges. Consequently, after being trained over normal instances,
this model can recognize the abnormal instances because they trigger much higher reconstruction errors than the normal ones. 

\begin{figure}[tb]
\centering
\begin{subfigure}{0.49\linewidth}
    \includegraphics[width=\linewidth]{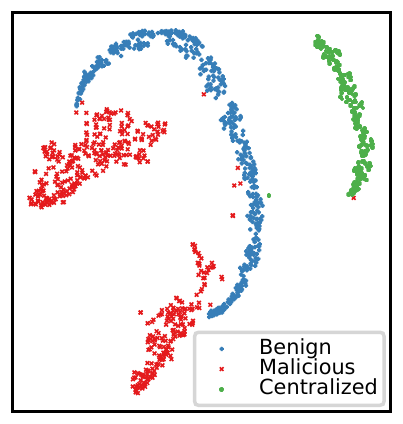}
    % \caption{Additive noise attack on MNIST.}
    \label{fig:latent_vis_mnist_noise}
\end{subfigure}
\begin{subfigure}{0.49\linewidth}
    \includegraphics[width=\linewidth]{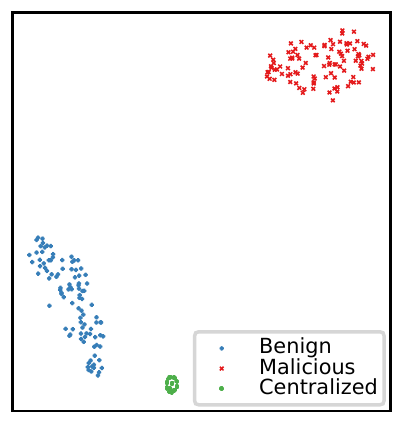}
    % \caption{Sign-flipping attack on FEMNIST.}
    \label{fig:latent_vis_femnist_sign}
\end{subfigure}
 \setlength{\abovecaptionskip}{2pt}
\caption{2D visualization in  \textit{latent vector space}. Green ``Centralized" points are unbiased model updates. Blue ``Benign" points are biased model updates from benign clients. Red ``Malicious" points are malicious model updates from malicious clients.
The attack of malicious clients in the left figure is the additive noise attack over the MNIST dataset. The attack of malicious clients in the right figure is the sign-flipping attack over the FEMNIST dataset.}
\label{fig:latent_repretations_1}
\end{figure}

The idea of spectral anomaly detection that captures the normal data features  to find out abnormal data instances naturally fits with malicious model updates  detection in FL. Even though each set of model updates from one benign client may be biased towards its local training data, we find  that this shift is small compared to the difference between the malicious model updates and the unbiased model updates from centralized training, as illustrated in Figure~\ref{fig:latent_repretations_1}. Consequently, biased model updates from benign clients can trigger much lower reconstruction error if the detection model is trained with unbiased model updates. Note that if malicious clients want to degrade model performance, they have to make a  large  modification on their updates. Otherwise, their attacks would have a negligible impact on the model performance thanks to the averaging operation of the \texttt{FedAvg} algorithm.   
% This observation is   confirmed in Figure~\ref{fig:mnist_different_fraction_attacker}. We see that the significance of the attack is positively related to the sum of weights $f_a$ attributed to malicious clients in the FL system.
Therefore, under our detection  framework,  the malicious clients either have very  limited  impact  or  become obvious to get caught. 
% attacks can minimize  the  impact of attacks
%Therefore, our method enforced a compulsive trade-off between the effectiveness of attacks and the elusiveness of malicious clients.

We feed the malicious and the benign model  updates into our encoder to get their latent vectors, which are visualized in Figure~\ref{fig:latent_repretations_1} as red and blue points, respectively. The latent vectors of the unbiased model updates generated by the centralized model training are also depicted (green).

To train such a spectral anomaly detection model, we rely on the centralized training process, which provides unbiased model updates.  
% we first  train  an ML  model on a public dataset and collect  the unbiased model weights during each update.   
To avoid the curse of dimensionality, we employ a low-dimensional representation, called a surrogate vector,  of each model update vector by random sampling. Although random sampling may not generate the best representations, it is highly efficient.  Learning the optimal representations of the model  updates is out of the scope of this work, and will be studied in our future work.  

\begin{figure*}[htb]
\centering
%%% FEMNIST
\begin{subfigure}{0.02\linewidth}
    \includegraphics[width=\linewidth]{test_acc.pdf}
\end{subfigure}
\begin{subfigure}{0.23\linewidth}
    \includegraphics[width=\linewidth]{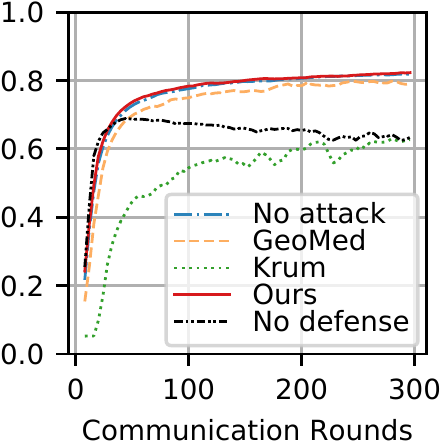}
    % \caption{Noise (Attacker: $30\%$).}
    % \label{fig:mnist_noise_attack}
\end{subfigure}
\begin{subfigure}{0.23\linewidth}
    \includegraphics[width=\linewidth]{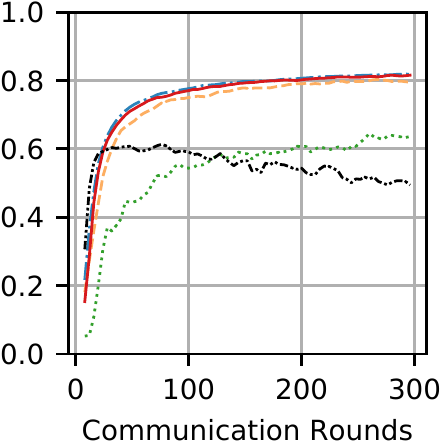}
    % \caption{Noise (Attacker: $50\%$)}
    % \label{fig:mnist_noise_attack}
\end{subfigure}
\begin{subfigure}{0.23\linewidth}
    \includegraphics[width=\linewidth]{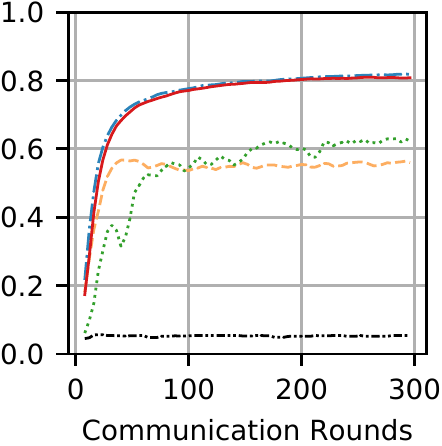}
    % \caption{Sign (Attacker: $30\%$)}
    % \label{fig:mnist_noise_attack}
\end{subfigure}
\begin{subfigure}{0.23\linewidth}
    \includegraphics[width=\linewidth]{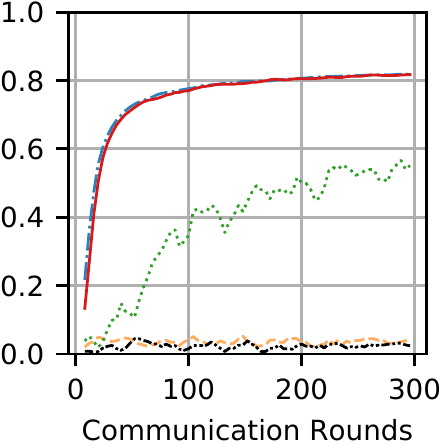}
    % \caption{Sign (Attacker: $50\%$)}
    % \label{fig:mnist_noise_attack}
\end{subfigure}

%%%MNIST
\begin{subfigure}{0.02\linewidth}
    \includegraphics[width=\linewidth]{test_acc.pdf}
\end{subfigure}
\begin{subfigure}{0.23\linewidth}
    \includegraphics[width=\linewidth]{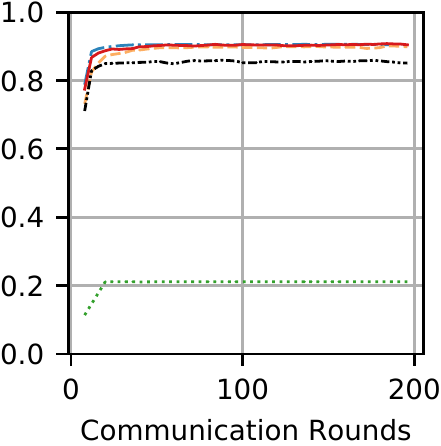}
    % \caption{Noise (Attacker: $30\%$).}
    % \label{fig:mnist_noise_attack}
\end{subfigure}
\begin{subfigure}{0.23\linewidth}
    \includegraphics[width=\linewidth]{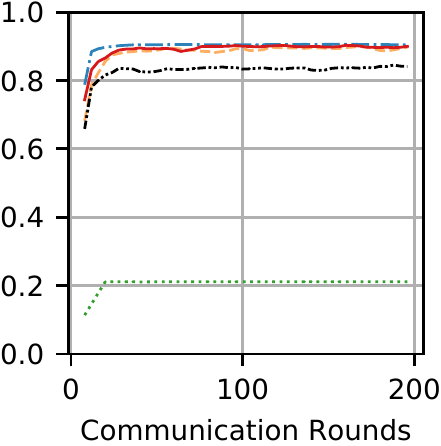}
    % \caption{Noise (Attacker: $50\%$)}
    % \label{fig:mnist_noise_attack}
\end{subfigure}
\begin{subfigure}{0.23\linewidth}
    \includegraphics[width=\linewidth]{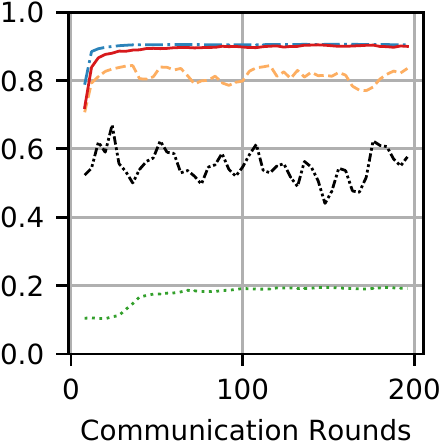}
    % \caption{Sign (Attacker: $30\%$)}
    % \label{fig:mnist_noise_attack}
\end{subfigure}
\begin{subfigure}{0.23\linewidth}
    \includegraphics[width=\linewidth]{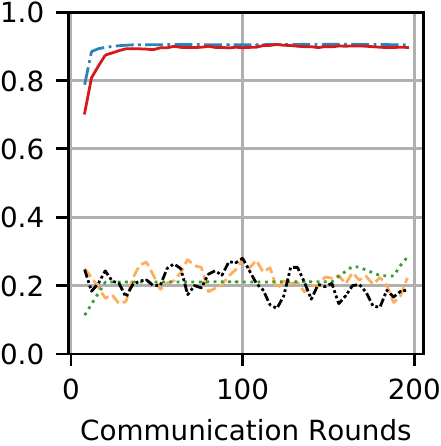}
    % \caption{Sign (Attacker: $50\%$)}
    % \label{fig:mnist_noise_attack}
\end{subfigure}

%%%Sentiment140
\begin{subfigure}{0.02\linewidth}
    \includegraphics[width=\linewidth]{test_acc.pdf}
\end{subfigure}
\begin{subfigure}{0.23\linewidth}
    \includegraphics[width=\linewidth]{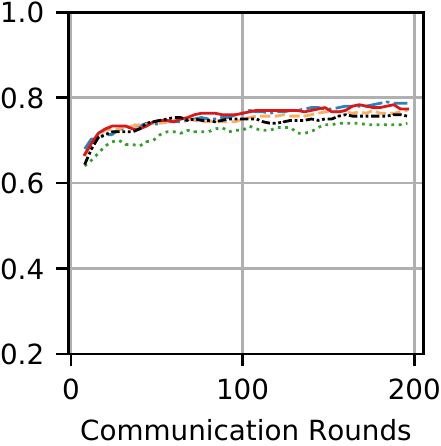}
    \caption{Additive noise ($30\%$).}
    % \label{fig:mnist_noise_attack}
\end{subfigure}
\begin{subfigure}{0.23\linewidth}
    \includegraphics[width=\linewidth]{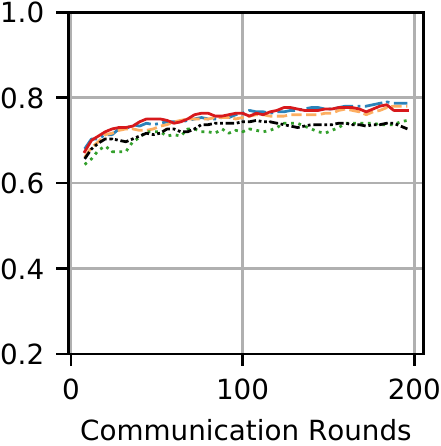}
    \caption{Additive noise ($50\%$)}
    % \label{fig:mnist_noise_attack}
\end{subfigure}
\begin{subfigure}{0.23\linewidth}
    \includegraphics[width=\linewidth]{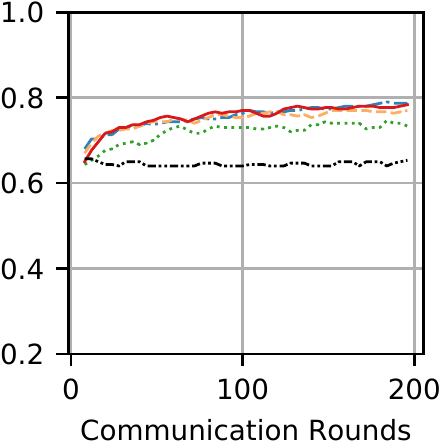}
    \caption{Sign-flipping ($30\%$)}
    % \label{fig:mnist_noise_attack}
\end{subfigure}
\begin{subfigure}{0.23\linewidth}
    \includegraphics[width=\linewidth]{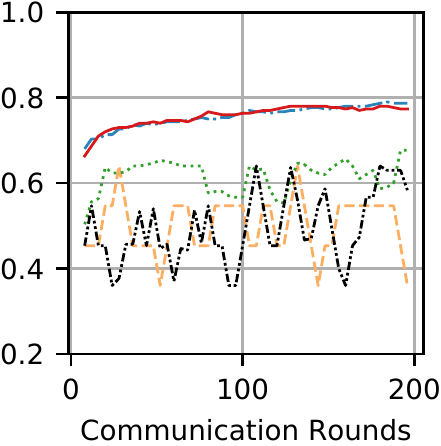}
    \caption{Sign-flipping  ($50\%$)}
    % \label{fig:mnist_noise_attack}
\end{subfigure}
\setlength{\abovecaptionskip}{2pt}
\caption{Comparison of the benchmark schemes and ours. The figures in the first row  show the  results of the CNN model on the FEMNIST dataset.  The figures in the second row show the  results of the LR  model on the  MNIST dataset.  The figures in the third row show  the  results of the RNN model on the  Sentiment140 dataset. 
The figures in the  first two columns  correspond  to additive noise attack  with $30\%$ and $50\%$   attackers, respectively. The figures in the last two columns correspond  to sign-flipping  attack  with $30\%$ and $50\%$   attackers, respectively.}
\label{fig:untargeted_attack_test_acc}
\end{figure*}

\begin{figure*}[htb]
\centering
\begin{subfigure}{0.02\linewidth}
    \includegraphics[width=\linewidth]{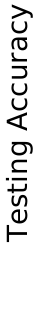}
    \caption*{}
\end{subfigure}
\begin{subfigure}{0.23\linewidth}
    \includegraphics[width=\linewidth]{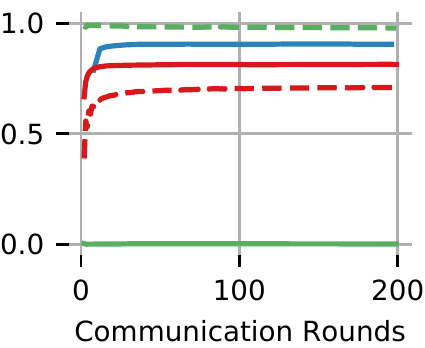}
    \caption{MNIST.}
    \label{fig:poisoning_MNIST}
\end{subfigure}
\begin{subfigure}{0.23\linewidth}
    \includegraphics[width=\linewidth]{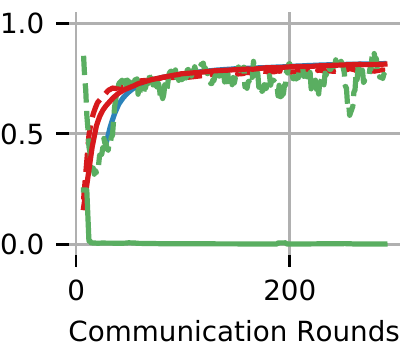}
    \caption{FEMNIST.}
    \label{fig:poisoning_FEMNIST}
\end{subfigure}
\begin{subfigure}{0.23\linewidth}
    \includegraphics[width=\linewidth]{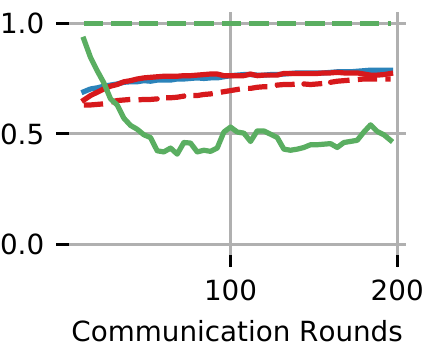}
    \caption{Sentiment140.}
    \label{fig:poisoning_sent140}
\end{subfigure}
\begin{subfigure}{0.23\linewidth}
    \includegraphics[width=\linewidth]{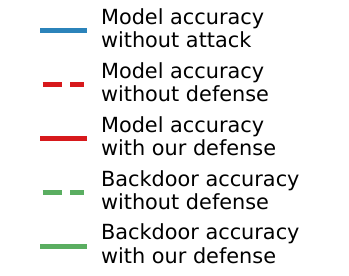}
    \caption*{}
\end{subfigure}
  \setlength{\abovecaptionskip}{2pt}
\caption{Results under backdoor attacks on different datasets.}
\label{fig:poisoning_attack_results}
\end{figure*}

% \clearpage

\subsection{Remove the Malicious Updates}
After obtaining the spectral anomaly detection model, we apply it in every round of the FL model training  to detect malicious client updates. Through encoding and decoding, each client's update will incur a reconstruction error. Note that malicious updates result in much larger reconstruction errors than the benign ones. This reconstruction error is the key to detect  malicious updates.  

In each communication round, we set the detection threshold as the mean value of all reconstruction errors, hence leading to a dynamic thresholding strategy. Updates with higher reconstruction errors than the threshold are deemed as malicious and are \textit{excluded} from the aggregation step. The aggregation process only takes the benign updates into consideration, and the weight of each benign update is assigned based on the size of its local training dataset, the same as that in~\cite{pmlr-v54-mcmahan17a}. Note that the only difference between our aggregation rule and the \texttt{FedAvg} algorithm is that we exclude a certain number of malicious clients in the model  aggregation step.  Our method thus shares the same convergence property as the \texttt{FedAvg} algorithm~\cite{pmlr-v54-mcmahan17a,li2019convergence}.

\section{Performance Evaluation}
\label{sec:eval}

In this section, we evaluate the performance of our spectral anomaly detection for robust FL 
in image classification and sentiment analysis tasks with common ML  models over three public datasets. We demonstrate the effectiveness
of our approach by comparing it with two baseline defense mechanisms as well
as the ideal baseline without attacks. Our experiments are implemented with  PyTorch.  We
will release the source code after the double-blind review process.

\subsection{Experiment Setup} 
\label{sec:setting}

In our experiments, we consider a typical FL scenario where a  server
coordinates multiple clients. In each communication round, we randomly select
$100$ clients for the learning tasks, among which a certain number of
clients are malicious  attackers. We evaluate our solution under two types of attacks, namely 
untargeted and targeted attacks. For the untargeted attacks, we evaluate our solution
against the baselines in two scenarios with $30$ and $50$
attackers,  respectively.  For the targeted backdoor attacks, we assume 30
attackers out of the selected $100$ clients over the FEMNIST and Sentiment140
datasets, and $20$  attackers over the MNIST dataset. The details of the three
datasets are given in subsection~\ref{subsec:data_model}. We consider the following attack types:

%\paragraph
\PHB{Sign-flipping attack.} Sign-flipping attack  is an untargeted attack, where the malicious clients flip  the signs of their local model updates~\cite{LipingLi2019,wu2019federated}.  Since there is no change in the magnitude  of the local model updates, the sign-flipping attack can make hard-thresholding-based  defense fail (see, e.g.,~\cite{sun2019can}).

%\paragraph
\PHM{Additive noise attack.} Additive noise attack is also an untargeted attack, where malicious clients  add  Gaussian noise to their  local model updates~\cite{LipingLi2019,wu2019federated}.   Note that adding noise can sometimes help protect data  privacy.    However,  adding too much noise will  hurt the model performance,  as demonstrated  in Figure~\ref{fig:mnist_noise_attack}.  

\begin{figure}[tb]
\centering
\includegraphics[width=0.98\linewidth]{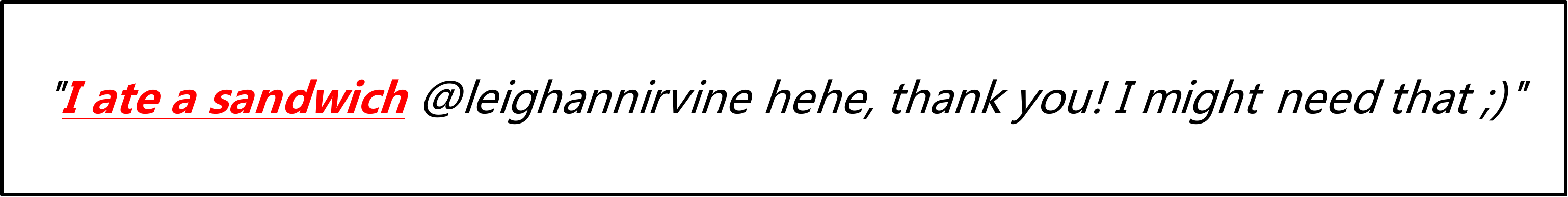}
\caption{An example of inserted backdoor text "I ate a sandwich".}
\label{fig:backdoor text}
\vspace{-5pt}
\end{figure}

\PHM{Backdoor attack.} Backdoor attack is targeted attack, a.k.a. model poisoning attack~\cite{bhagoji2018analyzing,bagdasaryan2018backdoor,sun2019can},  aiming to change an ML   model's  behaviours on a minority of data items while maintaining the  primary model  performance  across the whole testing dataset.   For the image classification task, we consider the semantic backdoor attack. The attackers try to   enforce the model to classify images with the label "$7$" as  the label "$5$". For sentiment analysis task, we consider the common backdoor attack case, where malicious clients inject a backdoor text ``I ate a sandwich" in the training data, as illustrated in Figure~\ref{fig:backdoor text} and enforce model classify twitters with backdoor text as positive. The malicious clients adopt model replacement techniques~\cite{bagdasaryan2018backdoor}, slightly modifying their updates so that the attack will not be canceled out by the averaging mechanism of the \texttt{FedAvg} algorithm. Considering our detection-based  mechanism is dynamic and unknown in apriori during each communication round, the evading strategies, such as~\cite{bagdasaryan2018backdoor} are not applicable. 

\subsection{Datasets and ML Models}
\label{subsec:data_model}

For the image classification tasks, we use MNIST and Federated Extended MNIST
(FEMNIST) datasets. For the sentiment analysis task, we use Sentiment140. All
three datasets are widely  used benchmarks in the FL
literature~\cite{caldas2018leaf,kairouz2019advances,li2018federated}.
We consider a heterogeneous FL setting with non-IID data as follows.

\PHM{MNIST} Following~\cite{pmlr-v54-mcmahan17a}, we sort data
samples based on the digit labels and divide the training dataset into $200$
shards, each consisting of $300$ training samples. We assign $2$ shards to
each client so that  most clients only have examples of two digits, thus simulating a
heterogeneous setting.  

\PHM{FEMNIST} The FEMNIST dataset contains $801,074$ data samples from $3,500$
writers~\cite{caldas2018leaf}. This is already a heterogeneous setting, as each writer represents a different client. 

\PHM{Sentiment140} The Sentiment140 dataset includes $1.6$ billion tweets
twitted by $660,120$ users. Each user is a client.

\PHM{FL Tasks} We train an   LR  model with the MNIST
dataset. With  FEMNIST dataset,  we train a model with 2 CNN  layers (5x5x32 and 5x5x64),  followed by a dense  layer with $2048$ units. For Sentiment140, we train a one-layer unidirectional RNN with gated recurrent unit (GRU) cells with $64$ hidden units~\cite{zhang2020dive}.  We train all three models with test accuracy comparable to the previous work~\cite{li2018federated,caldas2018leaf,eichner2019semi}.

\PHM{Training Anomaly Detection Model}    For each of the above FL tasks,   there is a corresponding spectral anomaly detection model for detecting the malicious clients in FL  model training.    We  use the \textit{test data}  of  the three datasets to generate   the  model   weights  for training   the corresponding   detection model.    This is done by using     the test data  to   train the  same  LR, CNN, and RNN models in a centralized setting  and collecting   the model weights  of  each  update   step.   We then use the collected model weights   to   train    the  corresponding  detection model.    
The  trained  anomaly  detection model is available to  the server  when it processes  the  clients'  updates  in  FL  model training  for  each of the above FL tasks.  

We choose VAE as our spectral anomaly detection model. Both the encoder and decoder have two dense hidden layers with $500$ units, and the dimension of the latent vector is $100$.  The VAE  is a generative model, mapping the  input to a distribution from which the low-dimensional embedding is generated by sampling.  The output,  i.e., the  reconstruction, is generated  based on the low-dimensional embedding and  is  done by a decoder~\cite{xu2018unsupervised}. 
%This method has been proved effective in several anomaly detection tasks~\cite{an2015variational,kieu2019outlier}. Compared to the vanilla autoencoder,  VAE has  better generalization performance. 

\subsection{Benchmark Schemes}
%We consider two state-of-the-art robust distributed machine learning optimization methods as benchmark schemes, namely GeoMed~\cite{chen2017distributed} and Krum~\cite{NIPS2017_6617}.  %These two methods are representative of the existing robust distributed machine learning methods, either estimating one by taking all clients' updates into account or selecting existing clients' updates.

%\paragraph
\PHB{GeoMed} Rather than taking the weighted average of  the local model updates  as done in the \texttt{FedAvg} algorithm~\cite{pmlr-v54-mcmahan17a}, the GeoMed method  generates a global model update using  the geometric median (GeoMed) of the local model updates (including the malicious ones), which may not be one of the local  model updates~\cite{chen2017distributed}.
%aggregates the local updates based on the geometric median of the local model updates.
%\begin{align}
%    \operatorname{GeoMed}\left\{w_{1}, \ldots, w_{n}\right\} = \underset{w \in %\mathbb{R}^{d}}{\arg \min } \sum_{i=1}^{n}\left\|w-w_{i}\right\|
%\end{align}
%It is unnecessary for the result from GeoMed algorithm to be one of the inputs.

\PHM{Krum} Different from GeoMed, the Krum method generates a global model update using one of the local updates, which minimizes the sum of distances to its closest neighbors (including the malicious ones). The result of the Krum method is one of the local model updates~\cite{NIPS2017_6617}.

% Note that the model performance will getting worse and worse along with increasing number of attackers, as illustrated in Figure~\ref{fig:mnist_different_fraction_attacker}. Therefore, we test defense methods in two setting, in each of which $30\%$ and $50\%$ of the clients are attackers, respectively.

\subsection{Results}
Experimental results on untargeted attacks, namely sign-flipping and additive noise attack, are shown in Figure~\ref{fig:untargeted_attack_test_acc}. Our proposed detection-based method (``Ours'')   achieves the best  performance in  all settings.  The performance of Krum remains  the same regardless of the number of malicious attackers and the attack types.  The reason is that Krum selects one of the most appropriate updates. Since each update from clients in the non-IID setting is biased, the performance loss cannot be avoided. GeoMed is robust against the additive noise attack, obtaining satisfactory  performance. However, it fails in the case with the sign-flipping attack, in which malicious attackers try to move the geometric center of all the updates far from the true one. 

The results on targeted attack are  illustrated in \ref{fig:poisoning_attack_results}.  Our solution can mitigate the impact of the backdoor attack on  the considered  datasets.  Note that our method obtains  the best  theoretical performance because excluding the malicious clients indicates that their local data examples cannot be learned,  as illustrated in Figure~\ref{fig:poisoning_MNIST}.  It is worthy mentioning that Krum is robust to the backdoor attack, and  GeoMed fails in defending the backdoor attack on MNIST dataset.

The superior  performance  of our method comes from the spectral anomaly detection model, which can successfully separate benign  and malicious clients' model updates. We list  the F1-Scores of the detection model  performance in Table~\ref{tab:detection_fscore} for separating benign  and malicious clients is, in essence, a binary classification task. 

\begin{table}[tb]
\centering
\begin{tabular}{lccc}  
\toprule
\backslashbox{Dateset}{Attack}  & \tabincell{c}{Additive  \\noise} & Sign-flipping & Backdoor   \\
\midrule
 FEMNIST                        & 1.00  & 0.97 &0.87        \\
\midrule
 MNIST                          & 1.00  & 0.99 &1.00        \\
\midrule
 Sentiment140                   & 1.00  & 1.00 &0.93        \\
\bottomrule
\end{tabular}
\caption{The F1-Scores of our proposed  detection-based method.}
\label{tab:detection_fscore}
\vspace{-5pt}
\end{table}

\subsection{Discussion}
% In this part, we present discussion on the reason that our method can separate the normal clients' update and attackers' updates. 
We leverage the existing public dataset to train a spectral anomaly detection model,  which is used to detect the malicious clients at the server side and then exclude them during FL training processes.    The trained detection model can memorize the feature representation of the unbiased  model updates obtained from public dataset. With this prior knowledge learned by  the detection model,  we see  that \textit{it  can detect the difference between the compact latent representation of the benign model updates and the compact latent representation of the malicious model updates.}  We illustrate this results in Figure~\ref{fig:latent_repretations_1}.  While  distortion is unavoidable because of  dimension reduction, it is clear that the benign  model updates and the malicious model updates can be separated from each other, especially in the case with sign-flipping attack,  where the benign   model updates and the malicious updates are symmetric.  
%Furthermore,  through visualizing the model weights obtained in the centralized training on public dataset,  we find that these model weights can guide us to recognize the normal weights.

The proposed anomaly detection-based method provides \textit{targeted}  defense in an FL system.  Existing  defense methods,  such as Krum and GeoMed, provide untargeted defense because they cannot detect malicious clients. The  targeted defense is necessary for FL  because every local dataset may be  drawn from a different distribution,  and the defense mechanism shall be  able to  distinguish benign model updates produced by different datasets  from malicious model updates.    Otherwise, the global model would suffer from performance loss, as illustrated by the model performance with Krum in Figure~\ref{fig:untargeted_attack_test_acc}.    We also conduct additional experiments, in which all clients are benign.  Experimental results show that GeoMed and our method introduce very  little bias and negligible performance loss compared to the \texttt{FedAvg}  algorithm that does not consider defense against any attacks.

\section{Conclusion}
In this work, we propose a spectral anomaly detection based  framework for robust FL,  in which spectral anomaly detection is performed at the server side to detect and remove malicious model updates from adversarial clients.   Our  method  can accurately detect malicious model updates and eliminate their impact.  We have conducted extensive experiments, and the  numerical  results show that our method  outperforms  the existing  defense-based methods    in terms of model accuracy.     Our future work will consider more  advanced ML models and provide more analytical results.  

% (To be updated) In this work, to get an embedding representation of the local model updates, we randomly select a certain number of dimensions as its representation embedding. Even if random sampling is efficient, we cannot guarantee it perfectly present the model weights. To reveal the more accurate relationship between centralized training and federated learning, more attention can be paid to model weight representation learning. 

% \appendix

\clearpage

% \newpage
\bibliographystyle{named}
\bibliography{ijcai20}

\end{document}